\documentclass[review]{elsarticle}
\usepackage{graphicx}
\usepackage{epstopdf}
\usepackage{subfigure}
\usepackage{bm}
\usepackage{amssymb}
\usepackage{color}
\usepackage{amsmath}
\usepackage{lineno,hyperref}
\begin{document}

\begin{frontmatter}

\title{Probing the Intra-Component Correlations within Fisher Vector for Material Classification\tnoteref{mytitlenote}}
\tnotetext[mytitlenote]{This work is supported by the Academy of Finland and Infotech Oulu}

%\author{Xiaopeng HONG\fnref{myfootnote}}[Oulu University]
\author[mymainaddress]{Xiaopeng Hong\corref{mycorrespondingauthor}}
\cortext[mycorrespondingauthor]{Corresponding author}
\ead{xiaopeng.hong@ee.oulu.fi}
\author[mymainaddress,mysecondaryaddress]{Xianbiao Qi}
\ead{qixianbiao@gmail.com}
\author[mymainaddress]{Guoying Zhao}
\ead{gyzhao@ee.oulu.fi}
\author[mymainaddress]{Matti~Pietik\"{a}inen}
\ead{mkp@ee.oulu.fi}

\address[mymainaddress]{Department of Computer Science and Engineering, University of Oulu, Finland}
\address[mysecondaryaddress]{Beijing University of Posts and Telecommunications}

%\maketitle

\begin{abstract}
Fisher vector (FV) has become a popular image representation. One notable underlying assumption of the FV framework is that local descriptors are well decorrelated within each cluster so that the covariance matrix for each Gaussian can be simplified to be diagonal. Though the FV usually relies on the Principal Component Analysis (PCA) to decorrelate local features, the PCA is applied to the entire training data and hence it only diagonalizes the \textit{universal} covariance matrix, rather than those w.r.t. the local components. As a result, the \textit{local decorrelation} assumption is usually not supported in practice.

To relax this assumption, this paper proposes a completed model of the Fisher vector, which is termed as the Completed Fisher vector (CFV). The CFV is a more general framework of the FV, since it encodes not only the variances but also the correlations of the whitened local descriptors. The CFV thus leads to improved discriminative power. We take the task of material categorization as an example and experimentally show that: 1) the CFV outperforms the FV under all parameter settings; 2) the CFV is robust to the changes of the number of components in the mixture; 3) even with a relatively small visual vocabulary the CFV still works well on two challenging datasets.

\end{abstract}


\end{frontmatter}

%------------------------------------------------------------------------- 
\section{Introduction}
\label{sec:intro}
Image representations, which extract informative cues from images, play a central role in various computer vision tasks such as material classification and object recognition. With the increasing demand for `ideal' image representations which are both highly discriminative in distinguishing instances from classes and well robust to the possibly large intra-class divergence in appearance, a number of image representations have been developed in the literature \cite{huang2014feature,chatfield2011devil}. 

The bag-of-visual words (BoW) model has become a popular framework for image representations. The BoW extracts the local features, which characterize the \textit{local} properties of images, and assigns them to the closed entries in a codebook. A codebook, more commonly known as a vocabulary~\cite{Ojala02,Lowe04,mikolajczyk2005performance,Varma05, Varma09,tan_tip10,bay2008speeded}, is usually generated by descriptor quantization, such as automatic vector quantization~\cite{Lowe04,mikolajczyk2005performance,Varma05, Varma09, ul2012lqp} and manual design~\cite{Ojala02,tan_tip10}.
The BoW uses the histogram of visual words as the image representation, to approximate the occurrence frequency of visual words (in the codebook) for a particular image.
In general, the BoW is robust against noise and background clutters. However, in order to make the vocabulary compact, the descriptor quantization usually learns or selects a rather small codebook (typically in a size of 200~to~1,000) to represent hundreds of thousands of local descriptors. It thus inevitably leads to considerable information loss~\cite{sanchez2013image,boiman2008defense}.

Recently, the BoW framework has been extended to the approaches which aggregate the local descriptors into a vector representation incorporating additional information~\cite{perronnin2007fisher, jegou2010aggregating, zhou2010image, picard2011improving,delhumeau2013revisiting}. 
The Fisher vector (FV)~\cite{chatfield2011devil,sanchez2013image, perronnin2007fisher, perronnin2010improving, perronnin2010large,  simonyan2013fisher, krapac2011modeling, peng2014action}, as one successful extension of the BoW, has aroused a great deal of research interest.
The FV models the visual words via a Gaussian mixture model (GMM). 
Given an image represented by a group of local descriptors, the FV measures the \color{black}{deviations} of local descriptors (from this image) w.r.t. the GMM parameters. 
Compared with the BoW, the FV brings additional information, thus leads to substantial improvement in accuracy. 
Moreover, usually a smaller vocabulary is required for the FV to achieve satisfying accuracy. 
In addition, the rich dimension of the FV makes the corresponding image representations more linearly separable. 
Hence the FV performs well with the efficient linear classifiers. 
The superiority of the FV has been shown in~\cite{huang2014feature, chatfield2011devil, peng2014bag}, where state-of-the-art feature aggregating approaches are empirically evaluated and the Fisher vector is suggested to be the most competitive one. 
Another example can also be found in~\cite{gosselin2013inria+}, in which the Fisher vector based systems obtain the best results in the FGCOMP fine-grained Challenge 2013.

\begin{figure}[t]
  \centering
  \subfigure[3D sample points from four-component GMM]{
    \label{fig:subfig:a} %% label for first subfigure
    \includegraphics[width=0.45\textwidth]{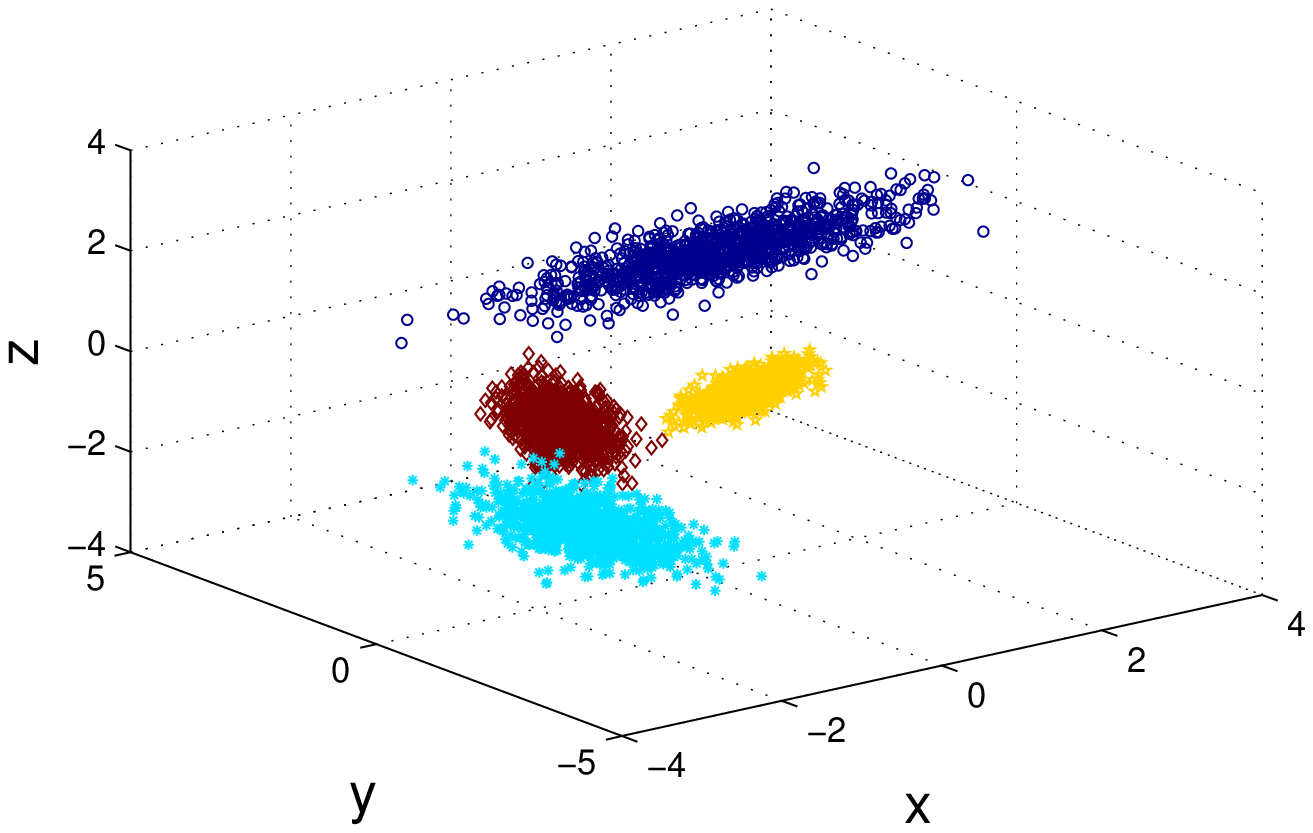}}
  %\hspace{1in}
  \subfigure[Projection to 2D sub-space by PCA]{
    \label{fig:subfig:b} %% label for second subfigure
    \includegraphics[width=0.45\textwidth]{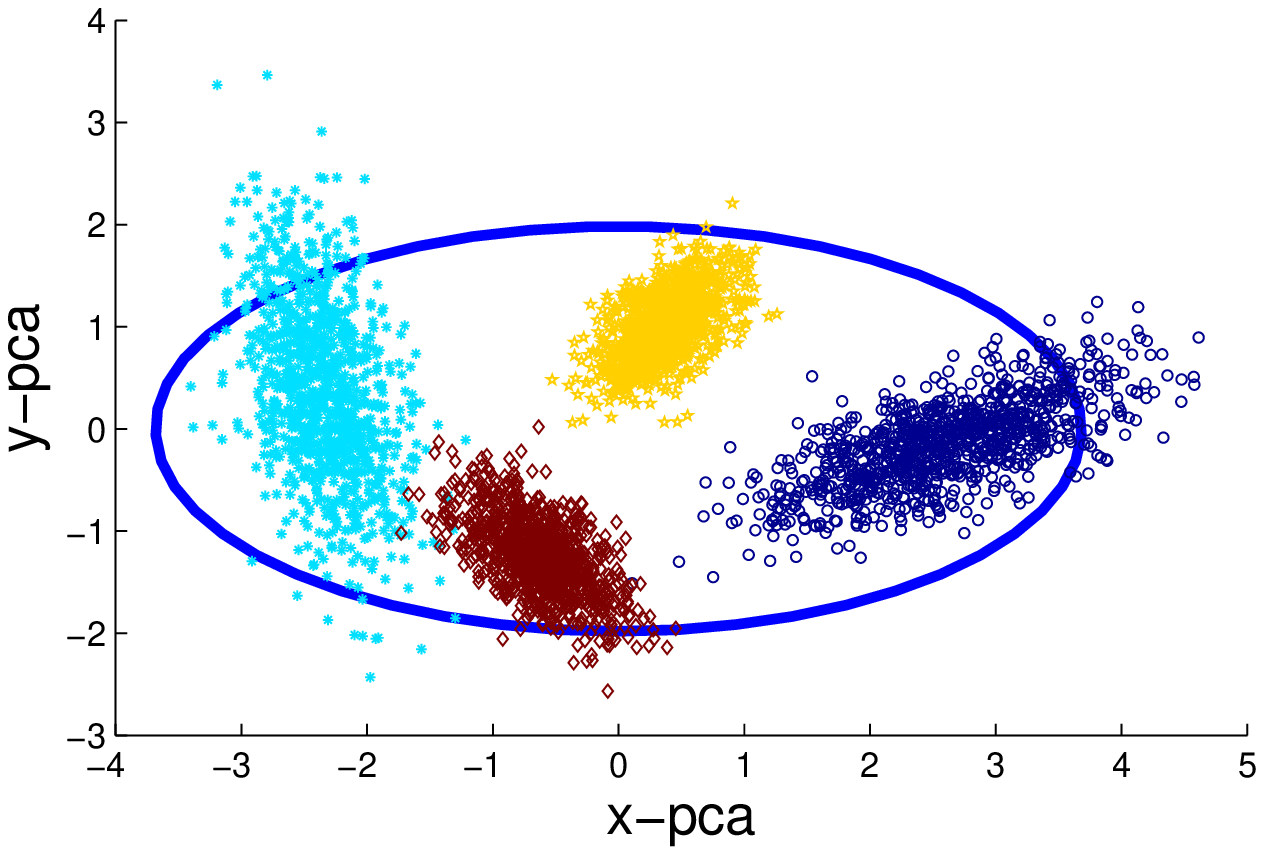}}
  \subfigure[The {\em universal} covariance matrix is diagonal]{
    \label{fig:subfig:c} %% label for third subfigure
    \includegraphics[width=0.45\textwidth]{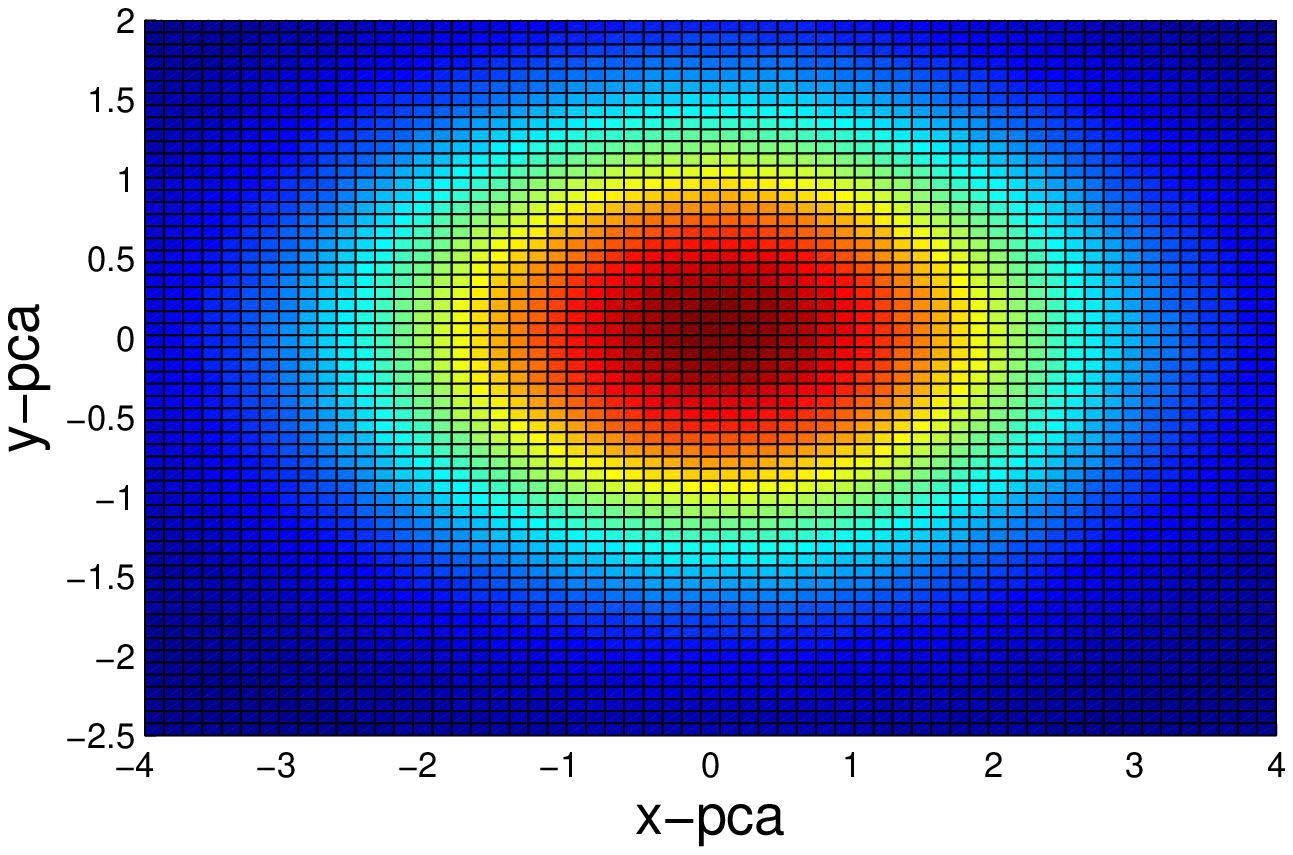}}
  \subfigure[\color{black}{Features are correlated within each component}]{
    \label{fig:subfig:d} %% label for fourth subfigure
    \includegraphics[width=0.45\textwidth]{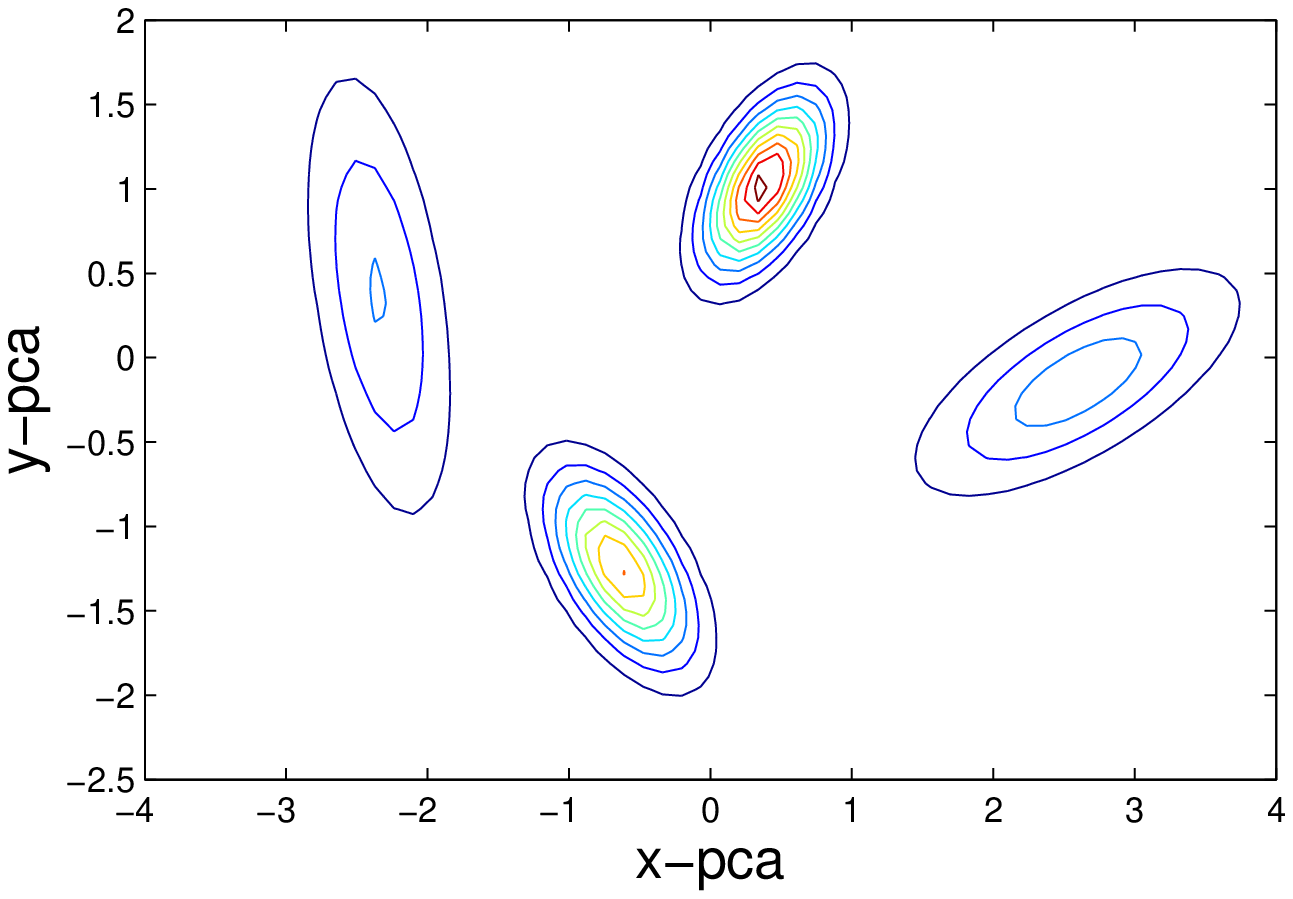}}        
  \caption{Universal PCA fails to eliminate the correlation within local modes}
  %\vspace{-10pt} 
  \label{fig:universal_pca} %% label for entire figure
\end{figure}
%\vspace{-10pt}

Despite of its great success, the existing Fisher vector framework is based on an underlying assumption that local descriptors are well decorrelated within each cluster (\textit{Local Decorrelation Assumption}). All covariance matrices of the components in the mixture are correspondingly simplified and forcedly restricted to be diagonal. Accompanied by several approximations, the assumption significantly eases the formulation of the FV~\cite{perronnin2010improving, simonyan2013fisher}. However, as a result, it severely restricts the FV to accurately characterizing the local features for images in complicated situations.

One might argue that the Fisher vector usually relies on the well-established Principal component analysis (PCA) to decorrelate local features and also reduce the dimensionality~\cite{simonyan2013fisher}. 
However, it should not be forgotten that the PCA is applied to the entire training samples. Thereby it is the {\em universal} covariance matrix (of the entire training samples) that the PCA only diagonalizes. There is no guarantee of the decorrelation for the local features from training images within each GMM component. Nor is it for those from testing ones.
%\vspace{-5pt}

Figure~\ref{fig:universal_pca} provides an example to illuminate the limitations of the \textit{Local Decorrelation Assumption} in the FV. 
Figure~\ref{fig:subfig:a} shows points sampled from a four-component GMM in the three-dimensional space. 
In Figure~\ref{fig:subfig:b}, they are projected to the 2D subspace by the PCA. 
Note that the PCA successfully decorrelates the entire set of samples and as a result diagonalizes the {\em universal} covariance matrix. 
It is demonstrated in Figure~\ref{fig:subfig:b} by the {\em non-sloping} blue ellipse, which stands for the contour line of the Gaussian probability density function on all points. Consistent observations can also be made from Figure~\ref{fig:subfig:c}, where the plotted surface of the Gaussian probability density with the {\em universal} mean and covariance matrix is not sloping either. However, the decorrelation assumption is not supported when taking each Gaussian component into account separately. 
In Figure~\ref{fig:subfig:d}, the apparent skewness of the contour line of each local Gaussian serves as a reminder of the undisputed existence of the feature correlations within each cluster. Obviously the \textit{local decorrelation} assumption is strictly satisfied in rare cases when the orientations of the principle axes for all GMM components happen to be close to those of the uni-modal Gaussian on all points. However in much more common cases it is violated if the orientations of the principle axes vary, as shown by the example here. A practical example in material classification is also provided to further illustrate this problem in Section \ref{subsec:cvd}.

Urged by the observation we reached, in this paper we propose to explore the feature correlations for improving the FV. We revisit the Fisher vector from the perspective of whitening transformation, and formally derive a vector representation taking into account feature correlations within the GMM clusters rather than only the variances (as the FV does). The resulting representation is named the Completed Fisher vector (CFV). It is a more general case of the FV where the \textit{local decorrelation} assumption is relaxed. In the following sections, we evaluate the proposed CFV and compare it with its counterpart namely the FV on two challenging data set  including KTH-TIPS2 and FMD. Superior performance on these two data sets successfully validate the effectiveness of the proposed CFV. 

The remaining of this paper is organized as follows: In Section 2, we derive our completed Fisher vector through in depth revisiting the standard Fisher vector. In Section 3.1, we provide a comparative study between the completed FV and the traditional diagonal ones on two common used local descriptors under different dimensions. The comparisons with the state of the art methods on two publicly available datasets are reported in Section 3. Finally, Section 4 concludes the whole paper.

%\vspace{-8pt}
\section{Exploring Intra-Component Correlations}
\label{sec:methodology}

In this section, the existing Fisher vector is firstly introduced to facilitate the derivation of the completed Fisher vector in Section~\ref{subsec:fv}. In Section~\ref{subsec:revistfv} we revisit the FV from the viewpoint of whitening transformation. Based on this understanding, we formulate the CFV and discuss on the implementation issues in Section~\ref{subsec:fcfv}.
%\vspace{-8pt}
\subsection{Standard Fisher Vector}
\label{subsec:fv}

The Fisher vector~\cite{perronnin2010improving} usually generates the vocabularies of visual words by means of the GMM. Let the set of parameters of a \textit{K}-mode GMM be
$\Theta = \left \{ \bm{\pi}, \bm{\mu}, \bm{\Sigma}\right \}$, where $\bm{\pi}\equiv\left \{\pi_1,\cdots ,\pi_K  \right \}$, $\bm{\mu}\equiv\left \{\bm{\mu}_1,\cdots ,\bm{\mu}_K  \right \}$, and $\bm{\Sigma}\equiv\left \{\bm{\Sigma}_1,\cdots ,\bm{\Sigma}_K\right \}$ are the sets of the priors, mean vectors, and covariance matrices of the \textit{K} components respectively. Denote by $I = \left \{\textbf{x}_n, n = 1\cdots N \right \}$ the set of \textit{D}-dimensional feature vectors extracted from an image. The GMM then associates each $\textbf{x}_n$ with the \textit{k-th} mode in the mixture via a soft assignment given by the posterior probability:
%\vspace{-6pt}
\begin{equation} \label{eq:softasg}
\lambda_{nk} = \frac {{\pi_k}\mathcal{N}\left ( \mathbf{x}_n, \bm{\mu}_k, \bm{\Sigma}_k\right )} {\sum_{t=1}^K {\pi_t}\mathcal{N}\left ( \mathbf{x}_n, \bm{\mu}_t, \bm{\Sigma}_t \right ) }, k = 1,\cdots,K, 
\end{equation}
where $\mathcal{N}\left( \mathbf{x}, \bm{\mu}_k, \bm{\Sigma}_k \right)$ for $k = 1,\cdots,K$ are the component Gaussian density functions of $\mathbf{x}$ with mean vector $\bm{\mu}_k$ and covariance matrix $\bm{\Sigma}_k$.

Under the \textit{local decorrelation} assumption, the covariance matrices of components are simplified to be diagonal, i.e. $\bm{\Sigma}_k \overset{\textup{s.t.}}=\bm{\Lambda}_k\equiv \textup{diag} \bm{\sigma}^2_k$,
 where $\bm{\sigma}_k \in \mathbb{R}_+^{D}$. The Fisher vector of image $I$ is denoted by $\mathcal{F}(I) = \left [ \cdots \mathbf{u}_k^T \cdots \mathbf{v}_k^T \cdots \right ]^T$, where $\mathbf{u}_k$ and $\mathbf{v}_k$ represent the first-order and second-order statistics of the deviations with respect to the parameters of the \textit{k-th} Gaussian~\cite{perronnin2007fisher, perronnin2010improving, simonyan2013fisher}. More specifically, the $d$-th entries of $\mathbf{u}_k$ and $\mathbf{v}_k$ for $d = 1,\cdots, D$ are defined as follows:
 %\vspace{-10pt}  
\begin{equation} \label{eq:ufv1d}
u_{dk} = {1 \over {N \sqrt{\pi_k}}} \sum_{n=1}^{N} \lambda_{nk} \frac{x_{dn} - \mu_{dk}}{\sigma_{dk}},
\end{equation}
 %\vspace{-8pt} 
\begin{equation} \label{eq:vfv1d}
v_{dk} = {1 \over {N \sqrt{2 \pi_k}}} \sum_{n=1}^{N} \lambda_{nk} \left[ \left(\frac{x_{dn} - \mu_{dk}}{\sigma_{dk}}\right)^2 - 1 \right].
\end{equation}

Let $\lambda_{k} = \sum_{n=1}^{N}\lambda_{nk}$ and re-parametrize the soft-assignment as $\bar{\lambda}_{nk} = \frac{\lambda_{nk}}{\lambda_{k}}$. Because $\bm{\Sigma}_k \overset{\textup{s.t.}}=\bm{\Lambda}_k$, Eqs.~\ref{eq:ufv1d} \textendash~\ref{eq:vfv1d} can be expressed by the following multivariate form:
%\vspace{-10pt}
\begin{equation} \label{eq:ufvmd}
\mathbf{u}_{k} \propto\lambda_{k}\sum_{n=1}^{N} \bar{\lambda}_{nk} \left [ \bm{\Lambda}^{-\frac{1}{2}}_k\left ( \mathbf{x}_n -\bm{\mu}_k \right ) \right ],
\end{equation}
%\vspace{-8pt}
\begin{equation} \label{eq:vfvmd}
\mathbf{v}_{k} \propto \lambda_{k}\sum_{n=1}^{N} \bar{\lambda}_{nk} \left\{ \left [\bm{\Lambda}^{-\frac{1}{2}}_k\left ( \mathbf{x}_n -\bm{\mu}_k \right )  \right ]\circ \left [\bm{\Lambda}^{-\frac{1}{2}}_k\left ( \mathbf{x}_n -\bm{\mu}_k \right )  \right ] - \mathbf{1} \right\},
\end{equation}
where $\mathbf{1}$ is the unit vector and the binary operator $\circ$ denotes the Hadamard product, a.k.a the entry-wise product.

The formal but involved derivations of the Fisher vector are provided in~\cite{perronnin2007fisher, simonyan2013fisher} from the viewpoint of the normalised gradients of the Fisher information matrix.
{\color{black}{We notice even under the \textit{local decorrelation} assumption, Eqs. \ref{eq:ufv1d} \textendash~\ref{eq:vfvmd} can only be obtained by approximating the Fisher information matrix and the partial derivatives w.r.t. its parameters.
In attempting to derive a more general form of the FV by incorporating the intra-component feature correlations, it would be inevitable to make more assumptions and approximations if along the same line of thought.
To avoid being enmeshed in such difficulties, we provide a second interpretation for the FV from the perspective of whitening transformation~\cite{duda2012pattern, jegou2012negative}, which provides a natural and more efficient way to introduce the corresponding formulas}}. 

%\footnote{In signal processing, the transform indicated by the \emph{A} part in Eq.~\ref{eq:ufv1d} includes the translating and whitening, which scale the axes such that the resulting variable $w_{d} = \frac{x_{d} - \mu_{dk}}{\sigma_{dk}}$ is towards zero mean and unit variance.}
%\vspace{-8pt}
\subsection{Fisher Vector: A Revisit}
\label{subsec:revistfv}

In probability theory and statistics, consider the \textit{D}-dimensional vectors in a set $I = \{\textbf{x}_n, n = 1\cdots N \}$ as discrete random variables. We have the expected value $\hat{\bm{\mu}}$ and covariance matrix $\hat{\bm{\Sigma}}$ as follows:
%\vspace{-10pt}  
\begin{equation} \label{eq:optmu}
\hat{\bm{\mu}} =E\left ( \mathbf{x} \right )=\sum_{n=1}^{N}p\left ( \mathbf{x}_n \right )\mathbf{x}_n,
\end{equation}
\vspace{-8pt} 
\begin{equation} \label{eq:optsigma}
\hat{\bm{\Sigma}} =
E\left (\left ( \mathbf{x}-E\left ( \mathbf{x} \right )  \right )\left ( \mathbf{x}-E\left ( \mathbf{x} \right )  \right)^T \right )=
\sum_{n=1}^{N}p\left ( \mathbf{x}_n \right )\left ( \mathbf{x}_n-\hat{\bm{\mu}}  \right )\left ( \mathbf{x}_n-\hat{\bm{\mu}}  \right)^T,
\end{equation}
where $p\left ( \mathbf{x} \right )$ is an underlying probability mass function (pmf) of $\mathbf{x}$, s.t. $p\left ( \mathbf{x} \right ) \geq 0$ and $\sum_{n=1}^{N}p\left ( \mathbf{x}_n \right ) = 1$ (\textit{the `Completeness' property}). 

Without losing generality, suppose $\hat{\bm{\Sigma}}$ to be positive definite. By left multiplying the inverse square root of $\hat{\bm{\Sigma}}$ on both sides of Eq.~\ref{eq:optmu}, and according to the distributivity of matrix multiplication we have
%\begin{equation} \label{eq:optmu2}
%\vspace{-8pt}  
\begin{align} \label{eq:optmu2}
\sum_{n=1}^{N}p\left ( \mathbf{x}_n \right )\hat{\bm{\Sigma}}^{-\frac{1}{2}}\mathbf{x}_n - \hat{\bm{\Sigma}}^{-\frac{1}{2}}\hat{\bm{\mu}}&=\sum_{n=1}^{N}p\left ( \mathbf{x}_n \right )\hat{\bm{\Sigma}}^{-\frac{1}{2}}\mathbf{x}_n - \sum_{n=1}^{N}p\left ( \mathbf{x}_n \right ) \hat{\bm{\Sigma}}^{-\frac{1}{2}}\hat{\bm{\mu}}&\\
\label{eq:optmu21}&=\sum_{n=1}^{N}p\left ( \mathbf{x}_n \right )\underset{\mathbf{w}_n}{\underbrace{ \left [\hat{\bm{\Sigma}}^{-\frac{1}{2}}\left (\mathbf{x}_n - \hat{\bm{\mu}}\right )\right ]}} =\mathbf{0}&,
\end{align}
%\end{equation}
\vspace{-8pt}

Similarly, by left and right multiplying $\hat{\bm{\Sigma}}^{-\frac{1}{2}}$ on both sides of Eq.~\ref{eq:optsigma}, the equation becomes:
\vspace{-8pt}  
\begin{equation} \label{eq:optsigma20}
\sum_{n=1}^{N}p\left ( \mathbf{x}_n \right )\hat{\bm{\Sigma}}^{-\frac{1}{2}}\left ( \mathbf{x}_n-\hat{\bm{\mu}}  \right )\left ( \mathbf{x}_n-\hat{\bm{\mu}}  \right)^T\hat{\bm{\Sigma}}^{-\frac{1}{2}} = \mathbf{I}.
\end{equation}
%\vspace{-5pt}  
Recall that $\hat{\bm{\Sigma}}^{-\frac{1}{2}}$ is symmetric such that $\hat{\bm{\Sigma}}^{-\frac{1}{2}} = [ \hat{\bm{\Sigma}}^{-\frac{1}{2}}]^T$, Eq.~\ref{eq:optsigma20} is equivalent to
%\vspace{-5pt}  
\begin{equation} \label{eq:optsigma21}
\sum_{n=1}^{N}p\left ( \mathbf{x}_n \right )\hat{\bm{\Sigma}}^{-\frac{1}{2}}\left ( \mathbf{x}_n-\hat{\bm{\mu}}  \right )\left (\hat{\bm{\Sigma}}^{-\frac{1}{2}}\left ( \mathbf{x}_n-\hat{\bm{\mu}}  \right)  \right )^T = \mathbf{I}.
\end{equation}

Taking into account the \textit{Completeness} constraint of \textit{pmf}, we rewrite Eq.~\ref{eq:optsigma21} by
%\vspace{-5pt}  
\begin{equation} \label{eq:optsigma22}
\sum_{n=1}^{N}p\left ( \mathbf{x}_n \right )\left [\hat{\bm{\Sigma}}^{-\frac{1}{2}}\left ( \mathbf{x}_n-\hat{\bm{\mu}}  \right )\left (\hat{\bm{\Sigma}}^{-\frac{1}{2}}\left ( \mathbf{x}_n-\hat{\bm{\mu}}  \right)  \right )^T - \mathbf{I}  \right ]=\mathbf{0}.
\end{equation}

By comparing Eqs.~\ref{eq:optmu21} and~\ref{eq:optsigma22} with Eqs.~\ref{eq:ufvmd} and~\ref{eq:vfvmd} respectively, we reach the following remarks:

1) It is well known that a linear transformation on the feature space will convert an arbitrary normal distribution into another normal distribution~\cite{duda2012pattern}. 
As revealed by Eqs.~\ref{eq:optmu21} and~\ref{eq:optsigma21}, an \textit{ideal} whitening process utilizing the expected value and the covariance matrix of $\left \{\mathbf{x}_n\right \}$, translates, scales, and rotates the axes so that 
the resulting multivariate variable $\mathbf{w}_n = \hat{\bm{\Sigma}}^{-\frac{1}{2}}\left ( \mathbf{x}_n-\hat{\bm{\mu}}  \right)\sim \mathcal{N}\left ( \mathbf{0}, \mathbf{I}\right )$, i.e. the standard Gaussian distribution.
However, in reality, for example in a GMM, the parameters are optimized on the local features from the entire training set rather than from a single image. 
It is thus difficult or even impossible for such pre-determined parameter settings to perfectly fit the expected mean and covariance matrix for a particular image $I$. 
As a result, the zero vector and the zero matrix on the right sides of Eqs.~\ref{eq:optmu21} and~\ref{eq:optsigma22} usually cannot be reached after whitening transformation. 

2) {\color{black}{The Fisher vector reflects the remainders as indicated by Eqs.~\ref{eq:ufvmd} and~\ref{eq:vfvmd}. It measures the misfit between the parameter settings of each Gaussian component and the set of local features from a particular image by aggregating the first and second order statistics of the differences between the whitened vectors, and the zero vector as well as the identity matrix. It is rational that the whitening transformation prevents certain features from dominating differences calculations merely because they have large numerical values}}.

3) Apparently, the whitening transformation in the FV is a special case of the general one as  it uses only diagonal covariance matrices. It simplifies the calculation, yet restricts the form of the probability density function and greatly limits the power to capture the important feature correlations from data. To remove these limitations, we propose to explore the intra-component correlations in the next section.
%\vspace{-8pt}
\subsection{Exploring Intra-Component Correlations}
\label{subsec:fcfv}

In this subsection we remove the \textit{local decorrelation} assumption and relax the constraint that the covariance matrix should be diagonal. In the light of the relation between Eqs.~\ref{eq:optmu21} and~\ref{eq:optsigma22} and Eqs.~\ref{eq:ufvmd} and~\ref{eq:vfvmd}, we define the derivations considering the feature correlations inside the GMM component as follows:
\vspace{-8pt}    
\begin{equation} \label{eq:ufullfv}
\mathbf{\tilde{u}}_{k} = {1 \over {N \sqrt{\pi_k}}} \sum_{n=1}^{N} \lambda_{nk} \left [ \bm{\Sigma}^{-\frac{1}{2}}_k\left ( \mathbf{x}_n -\bm{\mu}_k \right )\right ],
\end{equation}
%\vspace{-8pt} 
\begin{equation} \label{eq:vfullfv}
\mathbf{\tilde{V}}_{k} = {1 \over {N \sqrt{\pi_k}}} \sum_{n=1}^{N} \lambda_{nk} \left [ \bm{\Sigma}^{-\frac{1}{2}}_k\left ( \mathbf{x}_n -\bm{\mu}_k \right )\left (\bm{\Sigma}^{-\frac{1}{2}}_k  \left ( \mathbf{x}_n -\bm{\mu}_k \right )\right )^T - \mathbf{I}\right ].
\end{equation}

To form a vector representation, we unpack the entries in the matrix $\mathbf{\tilde{V}}_{k}$ into a vector $\mathbf{\tilde{v}}_{k}$. The number of entries in $\mathbf{\tilde{v}}_{k}$ can be reduced as the right side of Eq.~\ref{eq:vfullfv} is symmetric. In specific, we have $\mathbf{\tilde{v}}_{k} = \left [ \mathbf{{vd}}_{k}^T,\alpha\cdot \mathbf{{vt}}_{k}^T\right ]^T$, where $\mathbf{{vd}}_{k}$ is a $D\times1$ vector consisting of all diagonal entries of $\mathbf{\tilde{V}}_{k}$,
$\mathbf{{vt}}_{k}$ includes the $D\left ( D-1 \right )/2$ elements above the diagonal of $\mathbf{\tilde{V}}_{k}$, and $\alpha$ is a parameter which balances the contributions of $\mathbf{{vd}}_{k}$ and $\mathbf{{vt}}_{k}$. For each of the $K$ modes in the Gaussian mixtures, the whitened deviations to the standard normal distribution are completely specified by the $D+D\left ( D+1 \right )/2=D\left ( D+3 \right )/2$ parameters including the elements of the first order mean vector and the full second order covariance matrix.

As a result, the vector representation of image $I$ is the stacking of the vectors $\mathbf{\tilde{u}}_{k}$ and then of $\mathbf{\tilde{v}}_{k}$ for each of the $K$ modes in the Gaussian mixtures, i.e., $\tilde{\mathcal{F}}(I) = \left [ \cdots \tilde{\mathbf{u}}_k^T \cdots \tilde{\mathbf{v}}_k^T \cdots \right ]^T$.

In contrast with the FV, there is no assumption in the form of the component covariance matrix in the CFV. The second order part $\mathbf{\tilde{V}}_{k}$ is a $D \times D$ matrix covering both the variances and the correlations, while the one of the FV is a $D \times 1$ vector including only the variances, namely the entries in the diagonal of $\mathbf{\tilde{V}}_{k}$. The resulting vector representation $\tilde{\mathcal{F}}$ is therefore a completed model of the Fisher vector by the second statistics. It is named the Completed Fisher vector (CFV) in this paper. Obviously, the FV is a special case of the CFV. Only when the diagonal entries of $\mathbf{\tilde{V}}_{k}$ are used, the CFV degenerates to the FV.

Compared with the Fisher vector whose dimension is $2DK$, and the BoW whose dimension is only $K$, the CFV has a richer dimension of $D\left ( D+3 \right )K/2$. In general then, knowledge of the covariance matrix allows to reflect the dispersion of the data in any direction, or in any subspace. It thus contains significantly more information from the intra-component correlations. In addition, it can also be understood as an embedding of the local features in a much higher-dimensional space which is more separable for linear classification. 

In practice, before using the representation in a linear model (e.g. a support vector machine), we follow the two issues by the improved Fisher Vector (IFV)~\cite{perronnin2010improving, simonyan2013fisher}, including: 1) Power normalization, which applies the function $\left | z \right |^\gamma \textup{sign} \left ( z \right )$ to each entry of the vector, where $0\leqslant \gamma \leqslant 1$. It can be regarded as a post-processing procedure using Gamma correction. More in-depth study can be found in~\cite{chapelle1999support, vedaldi2012efficient}. 2) L2 Normalization, which implicitly discards the background information as suggested by~\cite{perronnin2010improving, simonyan2013fisher}.

To our best knowledge, there are two relevant works which more or less realize the limitations of the aforementioned \textit{Local Decorrelation Assumption}~\cite{picard2011improving,delhumeau2013revisiting}. Picard and Gosselin remove the GMM procedure by using the k-means clustering directly, and straightly use the second second order information to characterize the clusters~\cite{picard2011improving}. In~\cite{delhumeau2013revisiting}, Delhumeau et al. learn a local PCA rotation matrix for each cluster of the partitioned feature space. However, both two works are applied to the VLAD (namely the \textit{vector of locally aggregated descriptors}) framework rather than the FV, which is usually shown with better performance~\cite{chatfield2011devil}. As a result, for example, no whitening issue is discussed by their works, while in this paper, we derive our improved descriptor from the perspective of whitening. More details in the significant differences between the FV and the VLAD can be found in related publications~\cite{chatfield2011devil,jegou2010aggregating,perronnin2010improving}. Moreover, our work only uses the commonly used traditional PCA and thus does not rely on the time-consuming local PCA as \cite{delhumeau2013revisiting} does.

In the next section, we will experimentally show that additional information incorporated in the CFV brings improvement in terms of discriminative power.
%------------------------------------------------------------------------- 
%------------------------------------------------------------------------- 
%\vspace{-10pt} 
\section{Experiments}
\label{sec:exp}

We take the task of material categorization as an example to evaluate our descriptor.  As most modern methods have achieved high accuracy~\cite{sharan2013recognizing, cimpoi2013describing}, say over 96$\%$ on the databases such as KTH-TIPS~\cite{Caputo10}, CUReT~\cite{Dana99}, UIUC~\cite{lazebnik2005sparse}, and UMD~\cite{xu2009viewpoint}, which are shown to contain limited diversity of real-world material appearances, we choose two more challenging datasets:

\textbf{KTH-TIPS2a dataset} (KT2a)~\cite{Caputo05} contains four samples of eleven different materials, each at nine different scales and twelve different lightings and poses, totalling 4608 images. We follow the setup~\cite{Caputo05}, where three out of the four samples for each material are available for training, and the images of the remaining sample are left for testing. 

\textbf{Flickr Materials Database} (FMD)~\cite{sharan2009material} includes ten categories of materials, each of which includes one hundred color images. It was designed for capturing the appearance variations of real-world materials. One characteristic of FMD is the substantial intra-class variation as a result of the diverse selection of instances in each category. It is worth noting that even the human perception can only achieve an accuray of 84.9$\%$ on FMD~\cite{sharan2013recognizing}. Thus it is indubitably challenging. According to the protocol by Sharan et al.~\cite{sharan2009material}, for each category, we randomly choose 50 images for training and the remaining 50 for testing.

In this section, random experiments are carried out ten times and the average accuracy is reported as the final result on each dataset. Unless specified otherwise we use the one-versus-all SVM with the linear kernel based on LIBSVM~\cite{chang2011libsvm} with $C$=1. For the dense SIFT local descriptors and the Fisher vector, we use the implementation of Vlfeat~\cite{vedaldi08vlfeat}. The gmdistribtuion class in Statistics Toolbox of MATLAB is utilized to optimize the GMM parameters. The symmetric matrix square root is employed to realize $\bm{\Sigma}^{-\frac{1}{2}}$ for the CFV, and the only parameter of CFV, $\alpha$, is set to 0.25 cross all experiments.
%\vspace{-8pt} 
\subsection{Completed versus Diagonal}
\label{subsec:cvd}

In this subsection, we compare the completed Fisher vector with the FV on various numbers of GMM components. {\color{black}{As in~\cite{vedaldi08vlfeat}, all experiments start by doubling the resolution of the input image, and the local descriptors are densely sampled with a factor $\sqrt{2}$ between successive scales, resulting in seven scales from a ratio of 2.0 to 0.25. As the FV, we also use PCA to reduce the dimension of local descriptors for the CFV}}.

\begin{figure}[t]
  \centering 
  \subfigure[]{
  %\fbox{\rule{0pt}{1in} 
    \label{subfig:tips2adsift_d60} %% label for second subfigure
    \includegraphics[width=0.30\textwidth]{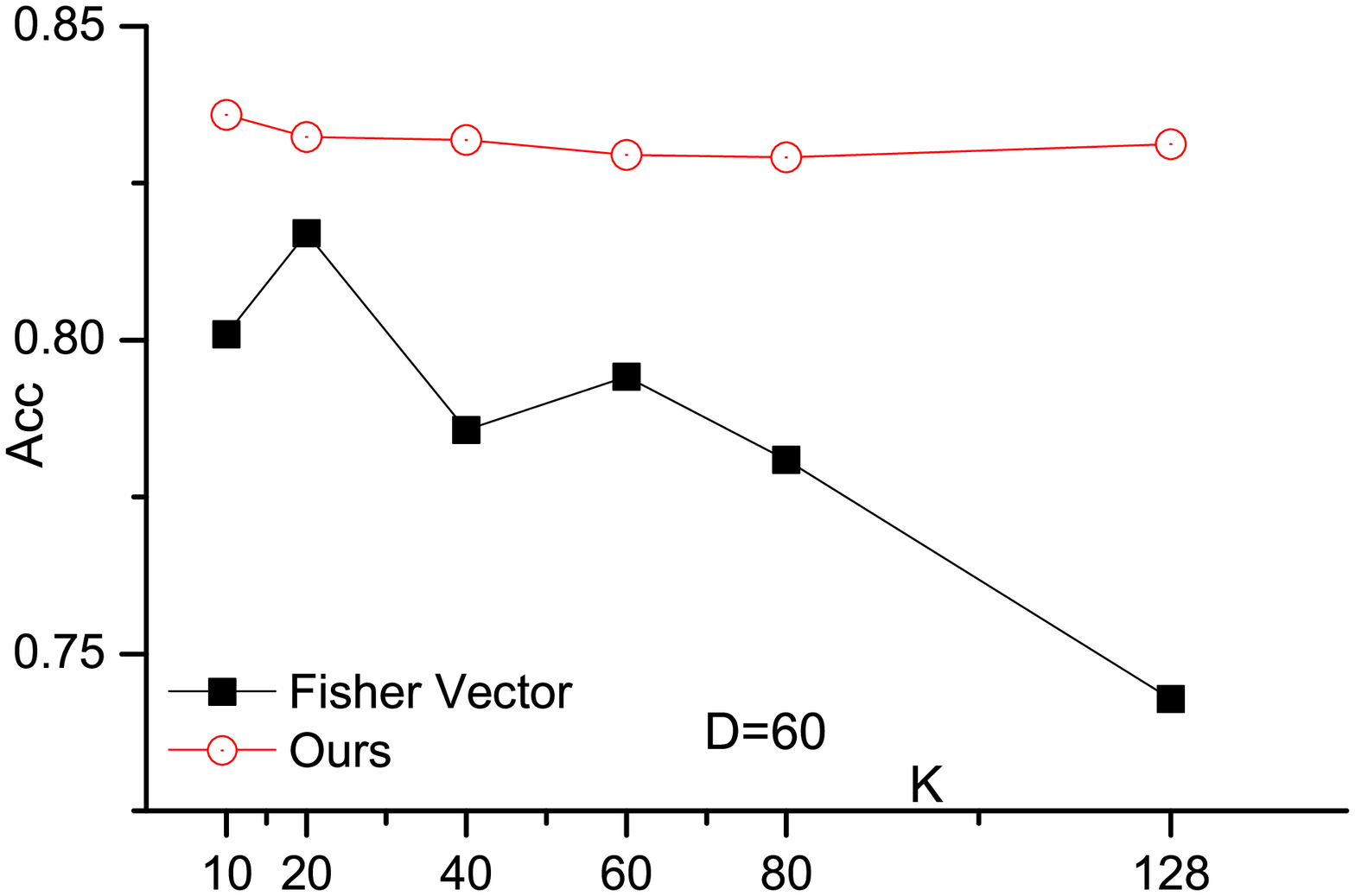}}
  \subfigure[]{
  %\fbox{\rule{0pt}{1in} 
    \label{subfig:tips2adsift_d80} %% label for third subfigure
    \includegraphics[width=0.30\textwidth]{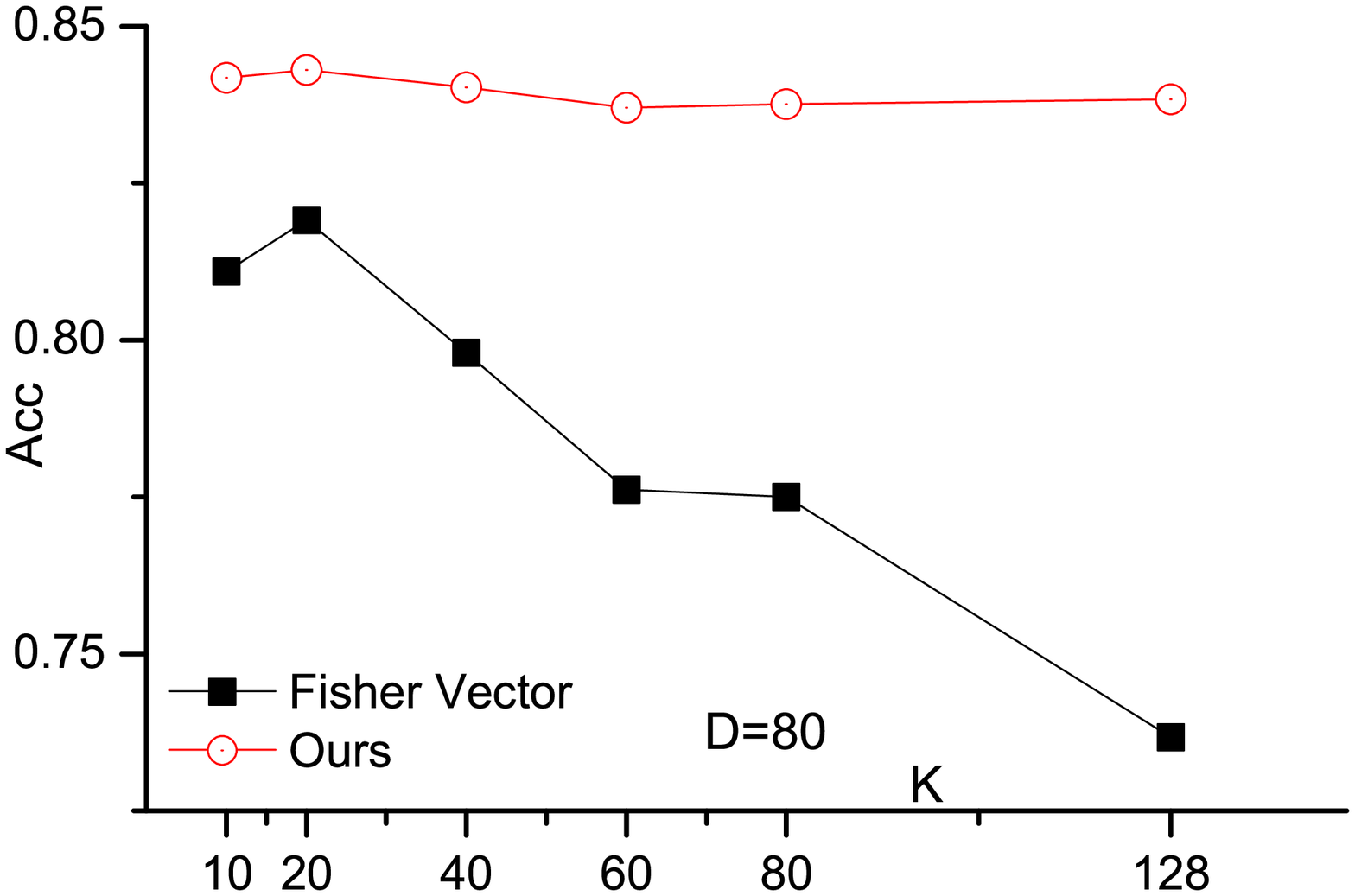}} 
  \subfigure[]{
  %\fbox{\rule{0pt}{1in} 
    \label{subfig:tips2adcollbp_d60} %% label for second subfigure
    \includegraphics[width=0.30\textwidth]{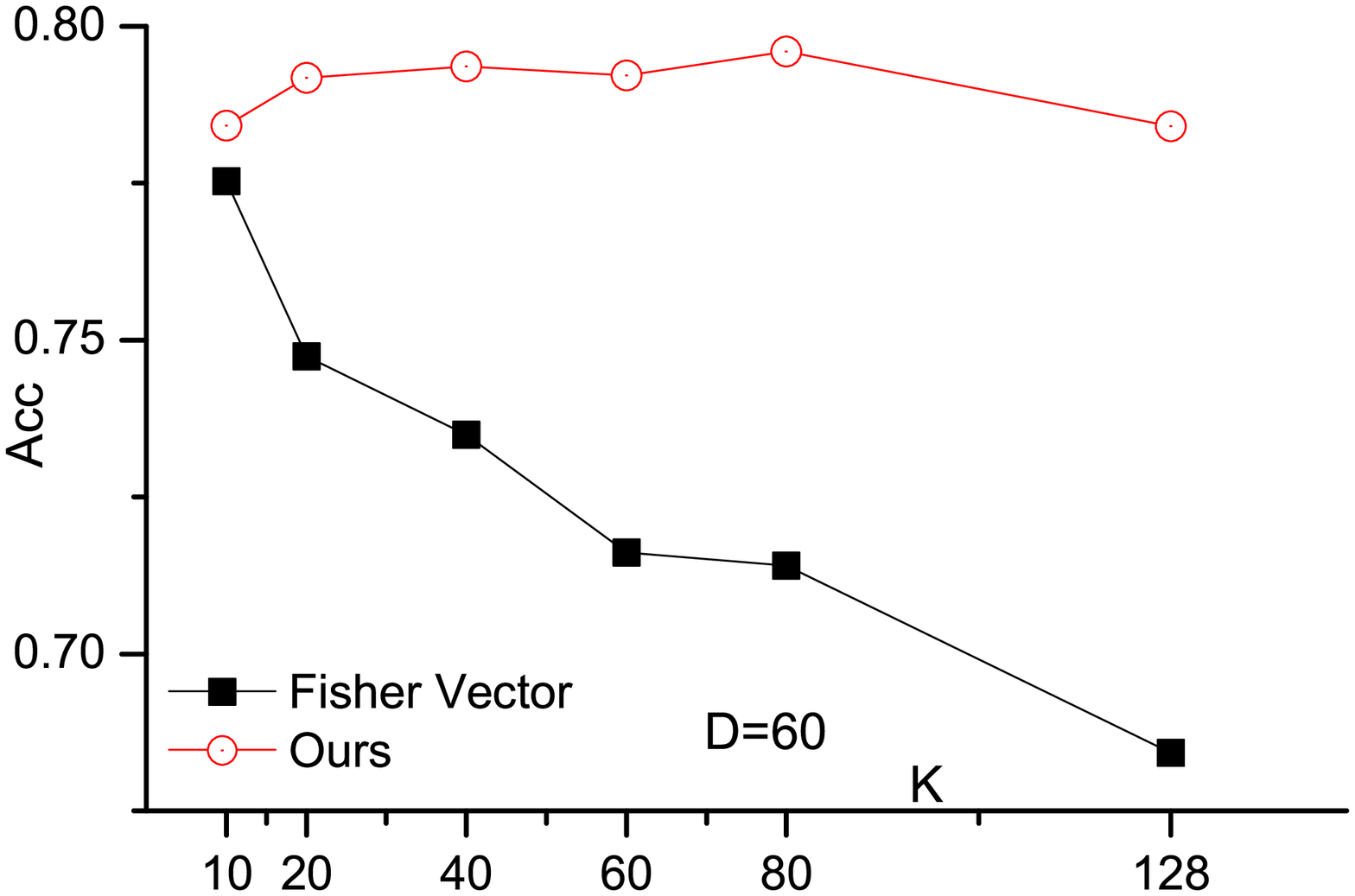}}
  \subfigure[]{
  %\fbox{\rule{0pt}{1in} 
    \label{subfig:tips2adcollbp_d80} %% label for third subfigure
    \includegraphics[width=0.30\textwidth]{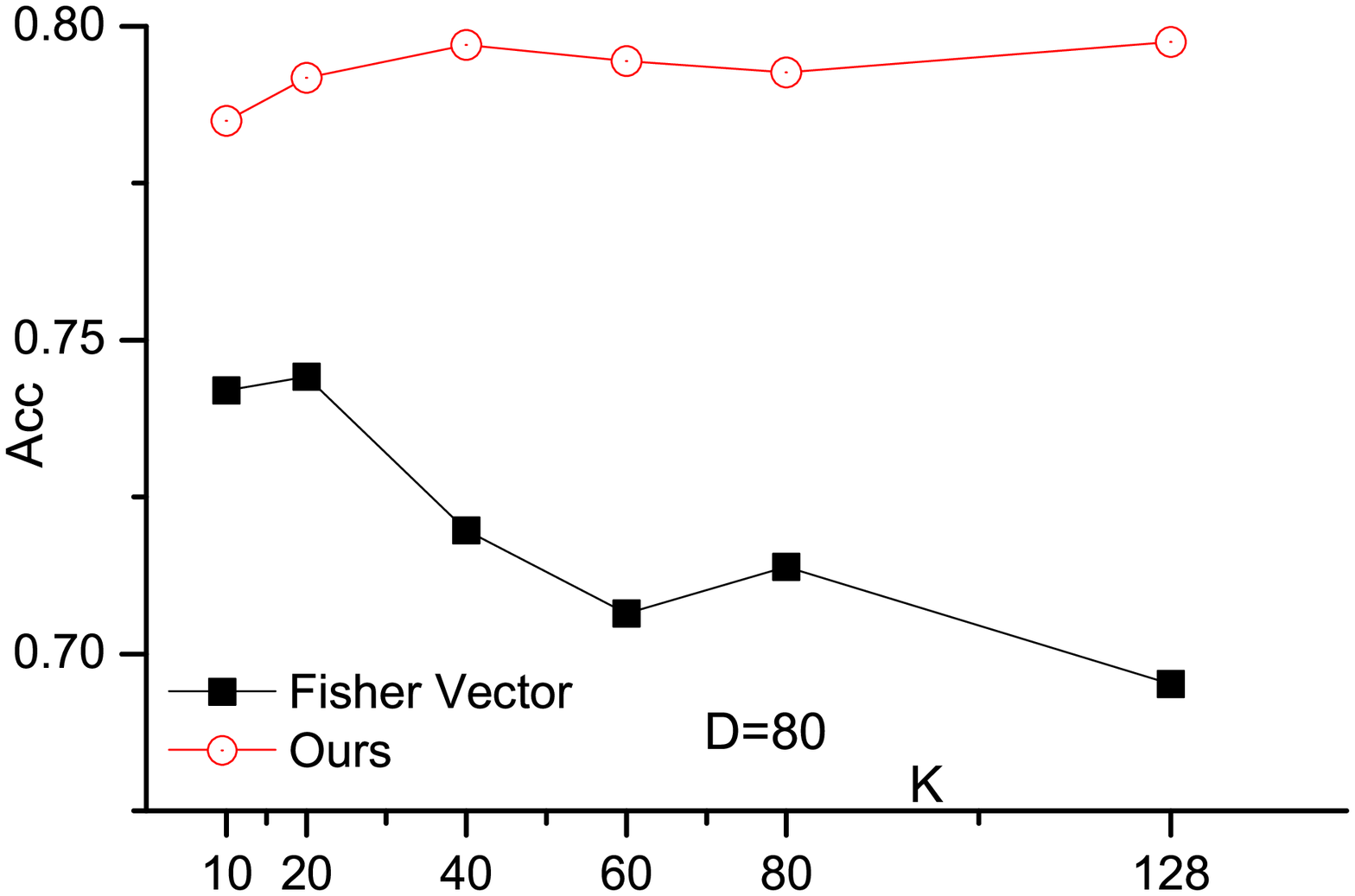}}      
  \subfigure[]{
  %\fbox{\rule{0pt}{1in} 
    \label{subfig:fmda} %% label for first subfigure
    \includegraphics[width=0.30\textwidth]{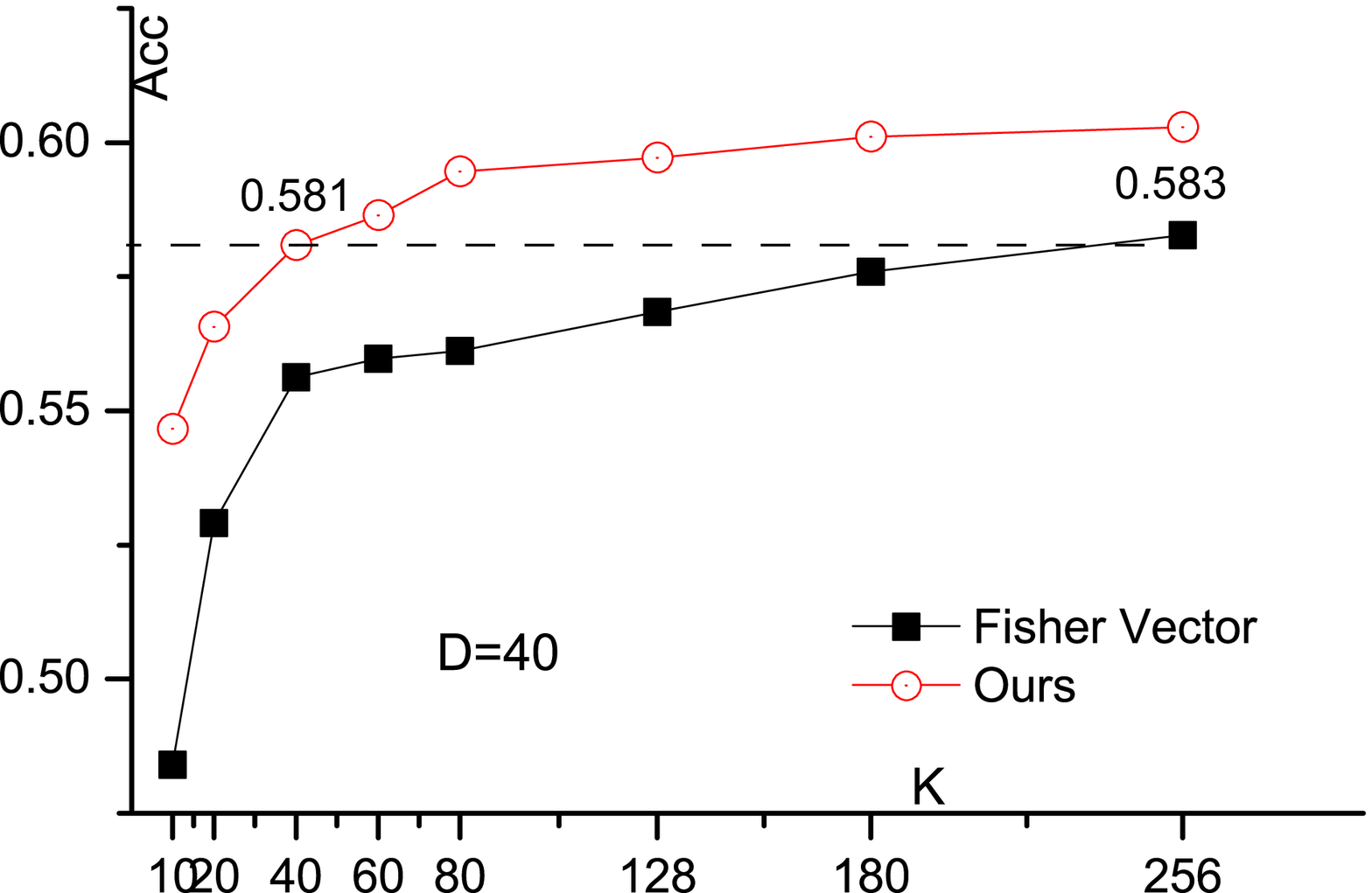}}
  \subfigure[]{
  %\fbox{\rule{0pt}{1in} 
    \label{subfig:fmdc} %% label for third subfigure
    \includegraphics[width=0.30\textwidth]{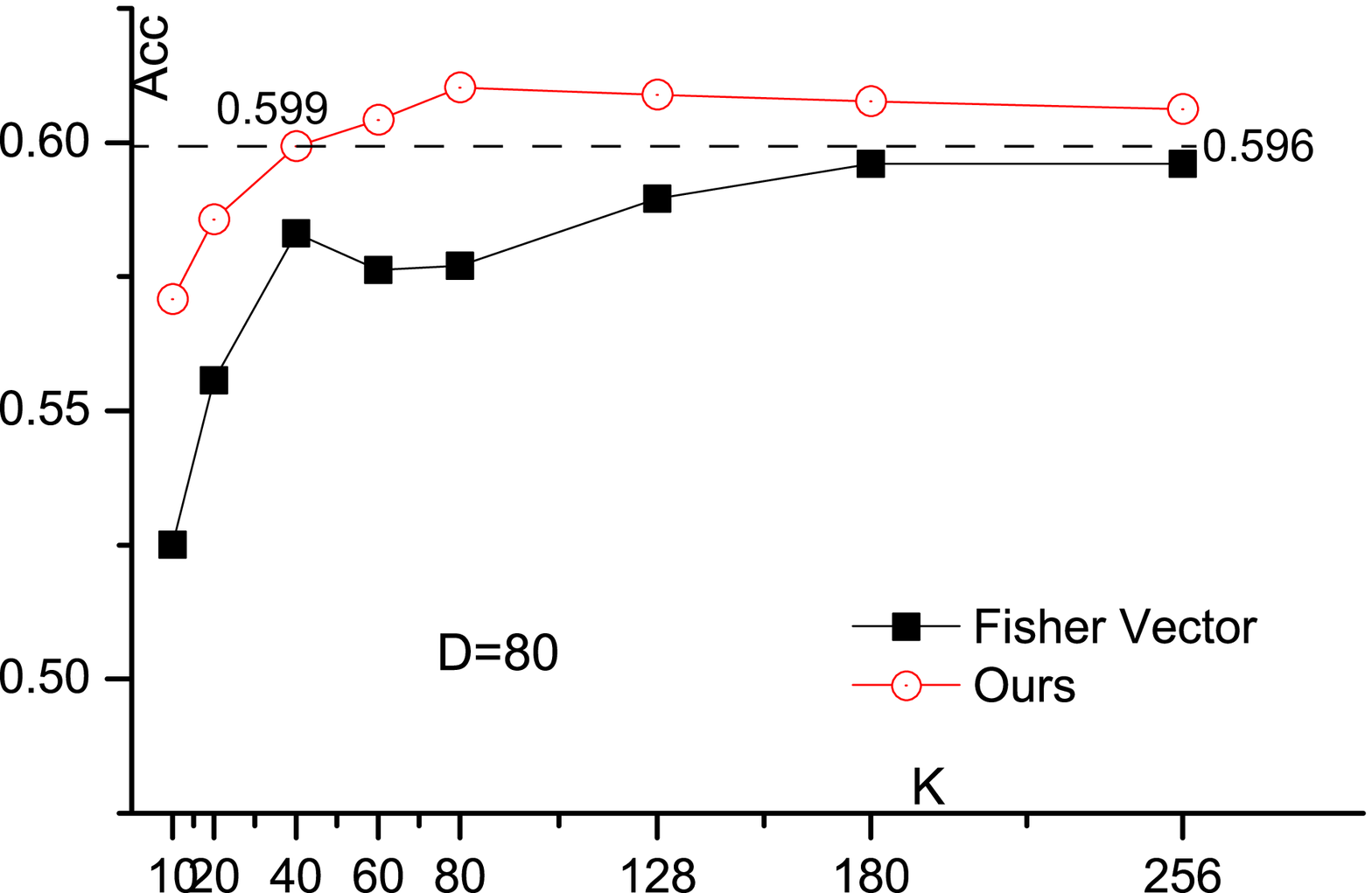}}  
  \caption{The comparisons between FV and CFV, (a-b) using dense SIFT as local descriptors on KT2a with D=60 and 80 respectively; (c-d)  using dense LBP on KT2a with D=60 and 80 respectively; (e-f) using dense SIFT on FMD with D=40 and 80 respectively.}
    \vspace{-10pt}  
  \label{fig:tips2adsift} %% label for entire figure
\end{figure}

On the KT2a dataset, we employ the following two local descriptors:

\textbf{Dense SIFT} captures a spatial histogram of local gradient orientations. We adopt the default settings of VLFeat~\cite{vedaldi08vlfeat}, i.e. the 128-dimensional dense SIFT with a step of four. The comparisons between our completed Fisher vector with the FV are illustrated by Figure~\ref{fig:tips2adsift}, where Figure~\ref{subfig:tips2adsift_d60} and~\ref{subfig:tips2adsift_d80} provide the results for $D=60$ and $80$ respectively. 

\textbf{Dense LBP} densely calculates the histograms of the 59-bin uniform LBP for a patch of size $16 \times 16$ in the RGB color channels separately. Then they are concatenated into a vector representation as the local descriptors. The comparative results are provided by Figure~\ref{subfig:tips2adcollbp_d60} and~\ref{subfig:tips2adcollbp_d80} for $D=60$ and $80$ respectively.

In line with the results, we have two important observations and related discussions:

1) The completed Fisher vector outperforms the FV under \textit{all} combinations of $K$, namely the numbers of the GMM components and $D$, the dimension of the descriptors after PCA regardless of the selection of local descriptors.

2) The KT2a dataset is special as for each material it only contains four samples. {\color{black}{Though there are 108 (or 72 for some samples) images under different lighting conditions and scales for each of the four samples, the modes required for fitting these variations are intrinsically limited. 
It can be observed that for both local descriptors, the optimal $K$ is around 20. 
The FV framework makes strong assumptions and approximations, which substantially affects the FV by how well the GMM models the data. Thus when $K$ is off the optimal value, the accuracy drops substantially, e.g., by about 10$\%$ when $K$ becomes 128. In contrast, as the completed Fisher vector relaxes nearly all assumptions which are adopted for deriving the diagonal FV, the CFV is much more robust to the quality of GMM estimates. As a result, the accuracy of the CFV remains almost the same even when $K$ is varying from 10 to 128.}}

We turn to the more challenging FMD dataset. Without losing generalization, we use the dense SIFT local descriptors. The comparison results are illustrated by Figure~\ref{subfig:fmda} and~\ref{subfig:fmdc} for $D=40$ and $80$ respectively. Obviously, more GMM components are required to fit the data since the intra-class variations on FMD are much larger than on KT2a. Nevertheless, the CFV still outperforms the FV for all combinations of $D$ and $K$, as in the case on KT2a. Moreover, the CFV is also more robust to the changes of $K$. As a result, by the setting of $K$ is 40, the CFV has achieved comparable or even better accuracy than the FV when $K = 256$. 

It is shown that the CFV is insensitive to the number of Gaussians, thus it is able to achieve an acceptable accuracy using much less components than the FV.
With a smaller visual vocabulary, the CFV is as efficient as, and even more efficient sometimes than the FV. The reason is that the number of components of GMM usually dominates the running time and thus the additional cost for calculating the feature coefficients turns to be trivial.

\begin{figure}[t]
  \centering 
  \subfigure[Distributions on training data]{
  %\fbox{\rule{0pt}{1in} 
    \label{subfig:coeff_train} %% label for second subfigure
    \includegraphics[width=0.45\textwidth]{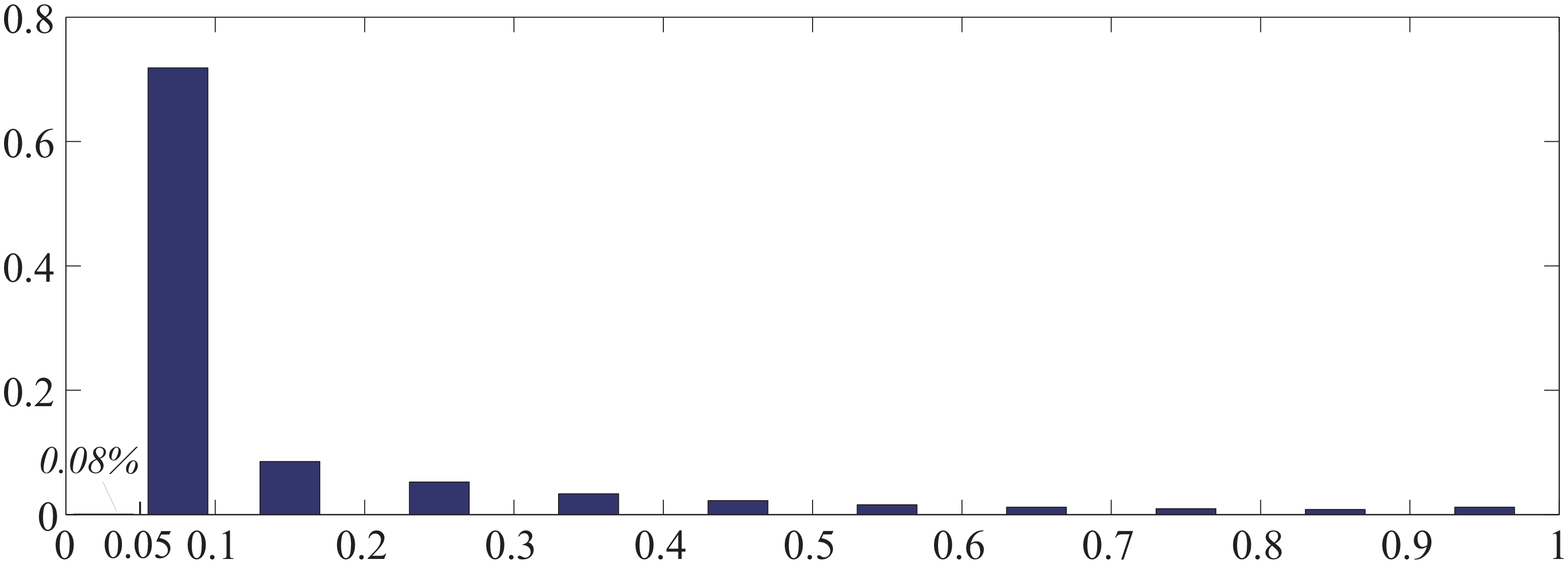}}
  \subfigure[Distributions on testing data]{
  %\fbox{\rule{0pt}{1in} 
    \label{subfig:coeff_test} %% label for third subfigure
    \includegraphics[width=0.45\textwidth]{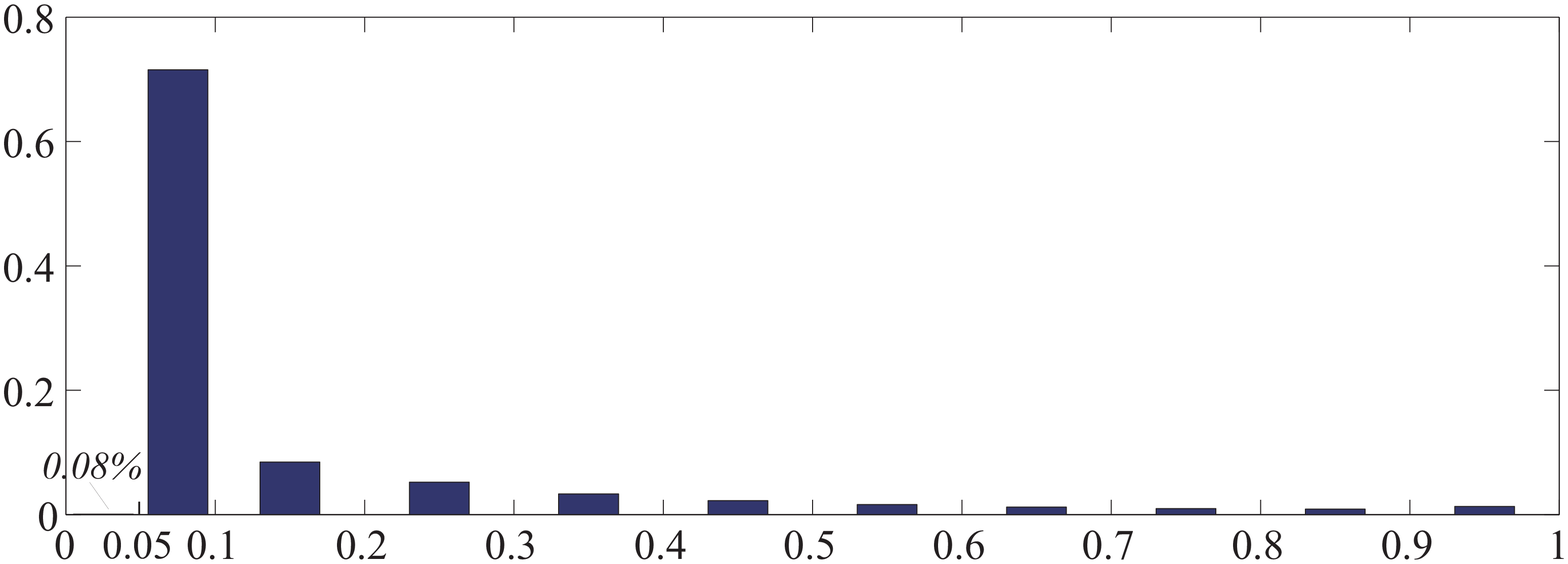}}  
  \caption{Distributions of correlation coefficients on FMD dataset}
  \vspace{-10pt} 
  \label{fig:fmd_mean_coeff} %% label for entire figure
\end{figure}

To further explain the rationality of the CFV, we report the average frequency of the Pearson correlation coefficients based on one particular split into training and test sets on the FMD dataset.
More specifically, when the parameter settings of the GMM ($D=80, K=256$) are determined, for the dense SIFT local features from each image, we calculated the Pearson correlation coefficients between any two of the feature entries using the soft assignment w.r.t each Gaussian given by Eq.~\ref{eq:softasg} as the probability mass function. 
The frequency of the absolute values of the correlation coefficients for one image is estimated via a histogram. The average histograms across all components of the GMM from all training and testing images are illustrated in Figure~\ref{subfig:coeff_train} and \ref{subfig:coeff_test} respectively. 
Notice that only a small portion, say 0.08$\%$ of the correlation coefficients is below an absolute value of 0.05 (where the two corresponding features are well uncorrelated). 
For the training data, 91.2$\%$ of the correlation coefficients are in the range of $[0.05, 0.5]$, while for the testing data, the portion in the same range becomes 90.8$\%$. 
The observation shows that the universal PCA fails to eliminate the feature correlations within the Gaussian components. It confirms our doubts about the underlined \textit{local decorrelation} assumption by the Fisher vector framework, and thus suggests the importance of exploring the intra-component feature correlation by the proposed CFV. 
%\vspace{-8pt}
\subsection{Comparisons with State of the Art}
\label{subsec:tips2}

We compare our approach with those state of the art on these two databases. The results of other methods are from corresponding publications. Tables~\ref{table:acc_tips2a} and~\ref{table:acc_fmd} list the accuracy.
It can be found that the CFV with the linear SVM achieves the best accuracy of 84.2$\%$ on KT2a and 61.2$\%$ on FMD among those methods under consideration. It is also worth noting that on the FMD dataset, the CFV performs as well as the current research hotspot, namely the deep convolutional neural network based approach~\cite{sermanet2013overfeat}, whereas on KT2a the mulsti-scale CNN representation~\cite{li2014learning}  doesn't show any superiority in performance. Since these two datasets are challenging, the promising results confirm that the feature correlations inside the components of the GMM are informative, and emphasize the superiority of the proposed CFV.

%\vspace{-15pt}
\begin{table}[!hbp] \small
\begin{center}
\setlength{\abovecaptionskip}{10pt}
\setlength{\belowcaptionskip}{0pt}
\caption{Performance on KTH-TIPS2a}
\label{table:acc_tips2a}
\begin{tabular}{|c|c|c|c|c|c|}
\hline
Approaches & Models &  Accuracy\\
\hline\hline
VZ-MR8~\cite{Caputo05} & $\chi^2$ SVM & 72.9\\
VZ-joint~\cite{Caputo05} &  $\chi^2$ SVM & 71.0 \\
3-scale LBP$^{riu2}$~\cite{Caputo05} &  $\chi^2$ SVM & 72.1 \\
4-scale LBP$^{riu2}$~\cite{Caputo05} &  $\chi^2$ SVM & 74.5 \\
LHS~\cite{sharma2012local} &  \textit{lin.}.SVM & 73.0\\
BOW~\cite{cimpoi2013describing} & \textit{lin.} SVM &74.8\\
 VLAD~\cite{cimpoi2013describing} & \textit{lin.} SVM &76.5\\
 IFV~\cite{cimpoi2013describing} & \textit{lin.} SVM & 82.5 \\
CCLBP~\cite{qi2013exploring} &  $\chi^2$ SVM & 77.4\\ 
multi-scale CNN~\cite{li2014learning} &  \textit{lin.} SVM & 77.4\\ 
\textbf{Ours} &  \textit{lin.} SVM & \textbf{84.2}\\

\hline
\end{tabular}
\end{center}
\end{table}
%\vspace{-20pt} 
\begin{table}[!hbp] \small
\begin{center}
\caption{Performance Comparison on FMD}
\label{table:acc_fmd}
\begin{tabular}{|c|c|c|c|c|c|}
\hline
Approaches & Models & Accuracy\\
\hline\hline
Eight-feature~\cite{sharan2013recognizing} & aLDA & 39.4\\
Eight-feature~\cite{sharan2013recognizing} & \textit{int.} SVM &55.6\\
Eight-feature w.t. mask~\cite{sharan2013recognizing} & aLDA & 42.0\\
Eight-feature w.t. mask~\cite{sharan2013recognizing} & \textit{int.} SVM &57.1 \\
BOW~\cite{vedaldi08vlfeat} & \textit{lin.} SVM & 46.0\\
VLAD~\cite{vedaldi08vlfeat} & \textit{lin.} SVM &  49.4\\
FV~\cite{vedaldi08vlfeat} & \textit{lin.} SVM &  59.6\\
\textbf{OverFeat.}~\cite{sermanet2013overfeat} & \textit{lin.} SVM &  \textbf{61.2}\\
\textbf{Ours} & \textit{lin.} SVM &  \textbf{61.2}\\
\hline
\end{tabular}
\end{center}
\end{table}
%\vspace{-15pt} 
%------------------------------------------------------------------------
%\vspace{-8pt}
\section{Conclusions}

This paper proposes a completed model of the Fisher vector to remove the limitations brought by the \textit{local decorrelation} assumption in the FV. As an image representation, the CFV encodes the completed statistics by the first two orders of the deviations of the whitened local descriptors w.r.t. the parameters of the GMM. 
It contains additional information compared with the FV, and thus leads to enhanced discriminative power. The CFV is more sophisticated than the FV, as the \textit{local decorrelation} assumption is relaxed. In addition, it is insensitive to changes of the number of the GMM components and thus it is able to achieve satisfying accuracy using a relatively small visual vocabulary. 

Like the FV, the CFV is efficient. The reduction in the size of vocabulary substantially decreases the running time. Even using the same size of vocabulary, the increase in running time for getting the additional information is moderate, as the inverse square root of the covariance matrix, the largest burden in calculating the CFV given a determined GMM, can be obtained prior to the evaluation procedure. Empirically, we only observe an increase by about 15$\%$ in the running time on FMD compared to the FV for $D=80$ and $K=256$. 

In future, we plan to combine the CFV with the deep convolutional neural network based approaches~\cite{Chatfield14} for further improvement.

\section{Acknowledgements}
This work was supported by the Academy of Finland and Infotech Oulu.

\section*{References}

\bibliography{fcfv_ref}
\end{document}